\documentclass[letterpaper,10pt,conference]{IEEEtran}
\IEEEoverridecommandlockouts

\usepackage{amsmath,amssymb}
\usepackage{booktabs}
\usepackage{graphicx}
\usepackage{microtype}
\usepackage{cite}
\usepackage{xcolor}
\usepackage{url}
\newcommand{\ind}{\mathrm{ind}}
\newcommand{\true}{\mathrm{true}}

\title{Marginal Calibration Does Not Compose:\\Hidden Dependence in Modular Robot Navigation}

\author{\IEEEauthorblockN{Rista Baral}
\IEEEauthorblockA{University of Delaware\\
\texttt{rbaral@udel.edu}}}
\begin{document}
\maketitle

\begin{abstract}
Robotic systems are typically composed of multiple independently developed modules that work together to perceive, predict, and act in the environment. Although each module may perform reliably in isolation, composing them does not necessarily preserve uncertainty calibration at the system level. In this work, we show that well-calibrated component interfaces do not necessarily produce calibrated downstream behavior after composition. Using a moving-obstacle prediction pipeline, we demonstrate that position and velocity estimators can each appear well calibrated individually, yet differences in how their error are correlated lead to substantially different estimates of future-state uncertainty. Consequently, assuming independence can make the system either overly confident or unnecessarily conservative, directly influencing downstream planning decisions and safety. Through simulations, we show that modeling the joint covariance restores downstream calibration and improves system performance, whereas dependence-robust uncertainty bounds enhance safety at the cost of increased conservatism. Our findings reveal a fundamental limitation of independently validating robotic modules and highlight the need for interfaces that communicate dependence information or support direct system-level calibration.

%Robotic modules are often validated separately, yet a robot acts on their composition. We show that calibrated marginal interfaces do not identify a calibrated downstream risk distribution. In a moving-obstacle pipeline, position and velocity estimators retain identical Gaussian marginals while only their error correlation changes. The composed future-state variance contains an unreported cross-covariance term; hence any point prediction rule using only the marginal interfaces fails for at least one compatible joint distribution. In 300,000 trials per condition, both components retain 95\% coverage, while independence-based future-state coverage ranges from 99.98\% to 86.46\%. In 30,000 matched closed-loop trials, ignored positive dependence produces 39.7 versus 8.7 collisions per 10,000; ignored negative dependence adds 4.2 stop steps per trial. Joint covariance estimation restores calibration and behavior. A dependence-robust bound prevents the positive-dependence failure but becomes conservative. The result defines a concrete interface requirement: marginal calibration evidence must be supplemented by dependence information, a validated contract, or direct system-level calibration.
\end{abstract}

\section{Introduction}
Modern robotic systems are built by combining multiple perception, estimation, prediction, planning, and control modules. This modular design has become a cornerstone of robotics because it allows individual components to be developed, tested, and improved independently. Many of these modules provide uncertainty estimates alongside their predictions, allowing downstream components to make safer and more informed decisions. In practice, these modules are typically validated individually before being integrated into the complete system.

It is therefore natural to assume that if every module accurately characterizes its own uncertainty, then the complete robotic system should also produce reliable uncertainty estimates. However, this assumption does not always hold. Although each module may be well calibrated individually, this does not guarantee calibrated system-level uncertainty after composition because module errors may be dependent. Ignoring these dependencies can make the composed system either overconfident or unnecessarily conservative, even when every individual module appears well calibrated in isolation.

This issue directly affects how robots make decisions. Many robotic systems rely on uncertainty estimates for probabilistic collision checking~\cite{dutoit2011}, risk-bounded planning~\cite{jasour2019,dawson2020}, and confidence-aware navigation~\cite{fridovich2020}. More broadly, recent work has shown that evaluating perception and prediction modules does not always reflect their impact on closed-loop system performance~\cite{ivanovic2022,corso2022,shen2024}. These observations raise a fundamental question: \emph{"Can independently validated uncertainty estimates from individual modules reliably describe the uncertainty of the composed robotic system?"}

To address this question, we study how uncertainty propagates when independently validated robotic modules are composed. Building on the theory of probabilistic calibration~\cite{gneiting2007}, we examine whether component-level calibration is preserved at the system level. Using a moving-obstacle prediction pipeline, we analyze how dependencies between module errors influence downstream uncertainty and examine the consequences of assuming independencies during composition.

\begin{figure*}[!t]
\centering
\includegraphics[width=0.99\textwidth]{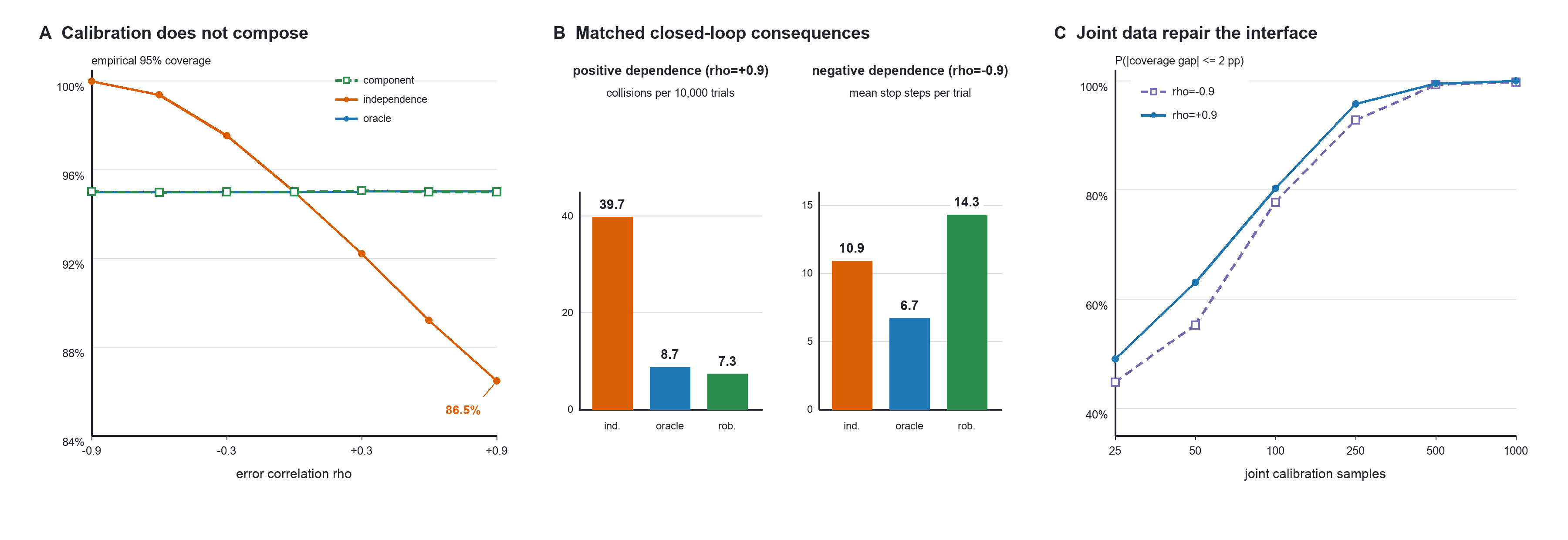}
\caption{Effect of error dependence on downstream calibration and robot behavior. (A) Component calibration remains unchanged, but downstream coverage varies under the independence assumption. (B) Ignoring error dependence changes closed-loop behavior. (C) Joint covariance estimation restores downstream calibration.}

\label{fig:results}
\end{figure*}

%\section{Calibration Is Not Identified by Marginals}
\section{Composing Calibrated Uncertainty Across Modules}
A central question in composing probabilistic robotic modules is whether the uncertainty guarantees provided by individual components remain valid after they are combined. To explore this question, consider a simple obstacle prediction problem. Let an obstacle have true position $p$ and velocity $v$, with future position after horizon $H$ given by $p_H = p + Hv$. Suppose two modules independently estimate position and velocity as

\begin{equation}
\hat p=p+\epsilon_p,\quad \hat v=v+\epsilon_v,
\end{equation}

where $\epsilon_p\sim\mathcal{N}(0,\sigma_p^2)$ and $\epsilon_v\sim\mathcal{N}(0,\sigma_v^2)$. Each module correctly captures its own uncertainty and is therefore individually calibrated. However, this marginal calibration does not specify the dependence between module errors. The joint error may therefore satisfy
\begin{equation}
\mathrm{Cov}(\epsilon_p,\epsilon_v)=\rho\sigma_p\sigma_v.
\end{equation}

After composition, the future position error becomes $\epsilon_H=\epsilon_p+H\epsilon_v$, with true uncertainty
\begin{equation}
\sigma^2_{\true,H}=\sigma_p^2+H^2\sigma_v^2+2H\rho\sigma_p\sigma_v. \label{eq:true}
\end{equation}

The cross-covariance captures the relationship between the two module errors and cannot be determined from the marginal uncertainties alone. In contrast, an interface that assumes independence yields

\begin{equation}
\sigma^2_{\ind,H}=\sigma_p^2+H^2\sigma_v^2. \label{eq:ind}
\end{equation}

\textbf{Proposition 1 (Non-identifiability of downstream calibration).}
Let $F_p,F_v$ be fixed calibrated marginals and $g$ a downstream map. If two couplings of $F_p,F_v$ induce different laws for $g$, no rule using only the marginals can be calibrated for both.

\emph{Proof.}
A marginal-only rule receives identical inputs under both couplings and returns the same law, which cannot equal two different pushforward laws. Equation~\eqref{eq:true} supplies a Gaussian witness whenever $H\sigma_p\sigma_v\neq0$ and $\rho_1\neq\rho_2$. \hfill$\square$

Thus, Gaussianity is not essential: for arbitrary marginals, an unspecified copula can leave the downstream law unidentified. The Gaussian case makes the calibration error analytically transparent.

This result shows that the missing information is not module uncertainty but error dependence. Assuming independence ($\rho=0$) ignores cross-covariance, causing overconfidence under positive correlation and under-confidence under negative correlation. Kalman filters and other joint-state estimators routinely retain such terms, and unknown correlation is classical in decentralized fusion~\cite{julier1997}. Our claim instead concerns separately developed or validated modules that expose only marginal uncertainty; the contribution is to quantify the calibration and closed-loop cost of that interface choice, not to propose a new fusion rule.

The impact of this independence assumption can be quantified analytically. When the composition rule reports a nominal central interval with Gaussian quantile $z_c$, the actual  coverage is
\begin{equation}
C_{\ind}(c,\rho)=2\Phi\!\left(z_c\frac{\sigma_{\ind,H}}{\sigma_{\true,H}}\right)-1. \label{eq:coverage}
\end{equation}

For the parameters used in our experiments, Eq.~\eqref{eq:coverage} predicts 99.98\% and 86.47\% coverage for a nominal 95\% interval when $\rho=-0.9$ and $\rho=0.9$, respectively, closely matching the empirical results. Thus, the simulations illustrate the practical consequences of an analytically characterized interface mismatch.

When the exact dependence is unavailable, an interface may instead report a bound $|\rho|\leq\bar{\rho}$. The compatible downstream variances are then bounded by
\begin{equation}
\sigma_p^2+H^2\sigma_v^2\ \pm\ 2|H|\bar\rho\sigma_p\sigma_v. \label{eq:budget}
\end{equation}
which provides a transparent and conservative uncertainty estimate without requiring a full joint model. Setting $\bar{\rho}=1$ recovers the robust bound in Eq.~\eqref{eq:robust}.

\begin{table*}[!t]
\centering
\caption{Representative results for three correlation settings. Future coverage evaluates downstream calibration, NLL measures predictive quality, and Brier score and ECE evaluate conflict-risk estimation. Collisions are reported per 10,000 closed-loop trials, and stop steps measure controller conservatism. Empirical composition estimates the joint covariance from 1,000 calibration samples.}
\label{tab:results}
\small
\begin{tabular}{ccrrrrrr}
\toprule
$\rho$ & Method & Future cov. & NLL & Brier & ECE & Coll. / 10k & Stop steps\\
\midrule
$-0.9$ & Independence & 99.98 & .659 & .077 & .073 & 1.7 & 10.89\\
       & Empirical    & 95.18 & .382 & .065 & .001 & 160.3 & 6.74\\
       & Oracle       & 94.98 & .382 & .065 & .001 & 164.0 & 6.74\\
       & Robust       & 100.00 & .891 & .090 & .112 & 0.0 & 14.33\\
\midrule
$0.0$  & Independence & 94.99 & 1.019 & .102 & .001 & 25.7 & 12.02\\
       & Empirical    & 95.84 & 1.021 & .102 & .006 & 21.7 & 12.21\\
       & Oracle       & 94.99 & 1.019 & .102 & .001 & 25.7 & 12.02\\
       & Robust       & 99.15 & 1.091 & .105 & .040 & 2.3 & 15.00\\
\midrule
$+0.9$ & Independence & 86.46 & 1.381 & .118 & .039 & 39.7 & 12.40\\
       & Empirical    & 95.25 & 1.292 & .115 & .001 & 7.7 & 15.38\\
       & Oracle       & 95.00 & 1.292 & .115 & .001 & 8.7 & 14.92\\
       & Robust       & 95.50 & 1.292 & .115 & .002 & 7.3 & 15.16\\
\bottomrule
\end{tabular}
\end{table*}

\section{Related Work and Positioning}

Existing robotic planning and navigation methods are based on probabilistic uncertainty estimates to evaluate collision risk, generate risk-aware plans, and provide safety guarantees ~\cite{dutoit2011,jasour2019,dawson2020}. These approaches generally assume that the uncertainty model required by the downstream planner is already available, rather than examining whether it can be recovered from uncertainty estimates produced by independently developed modules. As a result, the role of inter-module dependence in determining system-level uncertainty has received comparatively little attention.

%Existing robotic planning and navigation methods rely on probabilistic uncertainty estimates to evaluate collision risk, generate risk-aware plans, or provide safety guarantees~\cite{dutoit2011,jasour2019,dawson2020}. These approaches assume that the uncertainty model required for downstream decision making is already available. Our work instead considers a more fundamental system question: \emph{when independently validated modules are composed, is the information communicated through their interfaces sufficient to determine the downstream predictive uncertainty?} Proposition~1 shows that, in general, it is not.

A related challenge arises in distributed estimation, where the dependence between information sources is unknown. Conservative approaches such as covariance intersection address this challenge by producing consistent covariance estimates without requiring explicit correlation assumptions~\cite{julier1997}. Unlike covariance intersection, which addresses the fusion of estimates under unknown correlation, our focus is on the uncertainty information available at modular interfaces before fusion takes place.

%A related challenge arises in distributed estimation, where the dependence between information sources is unknown. Conservative methods such as covariance intersection avoid unsupported correlation assumptions by producing valid covariance estimates without requiring the true dependence structure~\cite{julier1997}. Unlike covariance intersection, which addresses fusion under unknown correlation, our focus is on the information available at modular interfaces before fusion takes place.

Recent work has shifted towards evaluating uncertainty based on its impact on downstream planning and robot behavior, rather than prediction accuracy alone~\cite{ivanovic2022,corso2022,shen2024,lidard2024}. Our work complements this line of work by examining the role of inter-module dependence while keeping each module's marginal uncertainty and calibration fixed. By isolating the effect of dependence, we show that it can substantially influence downstream uncertainty, task risk, and ultimately robot behavior.

Conformal methods construct calibrated trajectory regions for downstream planning~\cite{lindemann2023,dixit2023}. Their guarantees apply to the conformalized quantity under stated assumptions; they do not make separately calibrated module outputs jointly identified. End-to-end calibration of the planner-consumed output is therefore a complementary remedy.

\section{Controlled Evaluation}
To isolate the effect predicted by Proposition~1, all experiments keep the marginal uncertainty and module-level calibration fixed while varying only the dependence between position and velocity errors. Any difference in downstream uncertainty or robot behavior can therefore be attributed to the dependence structure.

\subsection{Matched uncertainty interfaces}
We set $\sigma_p=0.30$~m, $\sigma_v=0.20$~m/s, and prediction horizon $H=3$~s. For $\rho\in\{-0.9,-0.6,\ldots,0.9\}$, we jointly sample position and velocity errors while keeping both marginal distributions fixed. We compare four uncertainty interfaces: \emph{independence},~\eqref{eq:ind}; \emph{oracle},~\eqref{eq:true}; \emph{empirical}, which estimates the full $2\times2$ covariance from 1,000 calibration samples; and a \emph{robust} interface based on unknown-dependence bound,
\begin{equation}
\sigma^2_{\rm rob,H}=(\sigma_p+|H|\sigma_v)^2, \label{eq:robust}
\end{equation}
which follows from the Cauchy--Schwarz inequality.

Across all values of $\rho$, the position and velocity marginals, Gaussian forecasts, and nominal coverage of 50/80/90/95\% remain identical in expectation. Thus, the dependence between module errors is the only quantity that changes. Any differences in empirical coverage arise only from finite Monte Carlo sampling.

We evaluate each setting using 300,000 held-out samples. Downstream predictive calibration is measured using 95\% coverage and Gaussian negative log-likelihood (NLL). To evaluate task-level risk, predicted future means are sampled independently from $\mathcal U[-3,3]$~m, and the true future state is generated as $p_H=\hat p_H-\epsilon_H$. This makes the oracle predictive distribution exact while keeping the query distribution fixed. A conflict is defined as $|p_H|\leq0.5$~m. We report the binary Brier score and 15-bin expected calibration error (ECE). Because ECE depends on the choice of bins~\cite{arrieta2022}, we use it only as a descriptive reliability measure rather than the primary calibration metric.

\subsection{Closed-loop crossing controller}
Next, we evaluate uncertainty interfaces in a simple crossing scenario. A robot moves from $x=-3$~m to $x=3$~m at a constant speed of 1~m/s with $\Delta t=0.2$~s while obstacle crosses its path which starts at $y\sim\mathcal U[-4,-0.5]$~m with velocity $v_y\sim\mathcal U[0.5,1.0]$~m/s. At each control step ($\Delta t=0.2$~s), the robot predicts the obstacle's future position by composing noisy position and velocity estimates. It proceeds only if the predicted collision risk is below $\alpha=0.05$; otherwise, the robot stops and replans. Position and velocity errors remain correlated within each step but are independent across time. A collision occurs when $|x|,|y|\leq0.30$~m. For each correlation and uncertainty interface, we run 30,000 matched trials using the same simulated worlds and error realizations, recording collisions, stopping steps, and completion time. We also repeat the strongest positive and negative correlation settings using $\alpha\in{0.01,0.05,0.10}$ to evaluate the sensitivity to the planning threshold.

\subsection{Finite joint calibration}
Finally, we evaluate how much calibration data is needed to estimate the joint covariance. For
$N_{\rm cal}\in\{25,50,100,250,500,1000\}$,
we estimate the full covariance matrix 400 times at $\rho=\pm0.9$ using disjoint calibration and test sets. We report the fraction of trials whose analytical 95\% coverage error remains within a pre-specified two-percentage-point tolerance. All experiments use fixed random seed 20260719.

\section{Results}
\textbf{Component calibration does not guarantee system calibration.}
The position and velocity estimators remain individually well calibrated across all tested correlations, with their 95\% coverage staying within 0.06 percentage points of the nominal value. However, this agreement does not extend to the composed prediction. 
Under the independence assumption, downstream coverage varies from 99.98\% at $\rho=-0.9$ to 86.46\% at $\rho=+0.9$, whereas the oracle remains within 0.03 percentage points of the target 95\% coverage (Fig.~\ref{fig:results}A). The same trend is observed across all evaluated confidence levels. Proper scoring rules show the same behavior: at $\rho=+0.9$, the independence model has a higher Gaussian negative log-likelihood (1.381 vs.\ 1.292) and conflict Brier score (0.118 vs.\ 0.115) than the oracle (Table~\ref{tab:results}). Because the marginal uncertainty reported by both modules is identical in every experiment, these differences arise solely from the hidden dependence between their estimation errors.

\textbf{Miscalibrated uncertainty changes robot behavior.} The statistical mismatch is reflected directly in downstream decisions. The quality of the downstream risk estimates also degrades, with ECE increasing from 0.001 at $\rho=0$ to 0.039 at $\rho=+0.9$ and 0.073 at $\rho=-0.9$. Under positive dependence, ignoring correlation makes the planner overconfident, increasing collisions from 8.7 to 39.7 per 10,000 trials (4.6$\times$). Under negative dependence, the opposite occurs: the planner becomes overly conservative, reducing collisions to 1.7 per 10,000 at the cost of increasing average stopping time from 6.74 to 10.89 steps. Thus, lower collision rates arise from excessive caution rather than better-calibrated risk estimates.

\textbf{Modeling dependence restores calibration, while robust bounds trade efficiency for safety.} Incorporating the joint covariance restores downstream calibration, achieving future-state coverage of 95.18\% and 95.25\% for $\rho=-0.9$ and $\rho=+0.9$, closely matching the oracle. Even with only $N_{\rm cal}=250$ calibration samples, the estimated model remains within two percentage points of nominal coverage in over 92\% of repeated trials, increasing to nearly 100\% with 1,000 samples (Fig.~\ref{fig:results}). In contrast, the dependence-robust bound requires no joint calibration data and remains safe under positive dependence, but becomes overly conservative under negative dependence, producing 100\% coverage, an ECE of 0.112, and an average of 14.33 stop steps. The two approaches therefore represent different trade-offs between calibration accuracy and operational conservatism.

\section{Discussion}

Our results show that independently validating robotic modules is not sufficient to guarantee calibrated uncertainty after composition. Although each module may accurately quantify its own uncertainty, the downstream prediction also depends on how the module errors are related. Consequently, two systems can expose identical module-level uncertainty while producing different downstream uncertainty estimates and different robot behavior.

This finding has practical implications for modern robotic systems, which increasingly combine independently developed modules. Reporting only marginal uncertainty is therefore insufficient for downstream components to make reliable uncertainty-aware decisions. Instead, our results suggest that uncertainty interfaces should communicate dependence information when available, provide conservative uncertainty bounds when dependence is unknown, or alternatively calibrate uncertainty at the level of composed system.

Our controlled two-module Gaussian study establishes mechanism, not empirical behavior in nonlinear systems. In a pipeline such as an obstacle detector feeding a trajectory transformer, shared inputs may create context-dependent, multimodal coupling, while estimating the full joint density is impractical. A realistic audit can log synchronized residuals and calibrate a task-relevant projection, trajectory score, or collision-risk output, conditioned on operating regime and rechecked under shift. Without joint data, set-valued or dependence-robust interfaces remain alternatives. These are implications of Proposition~1, not effects demonstrated by our experiment.

\section{Conclusion}
We showed that independently calibrated robotic modules do not necessarily produce calibrated uncertainty after composition. Although two systems may expose identical module-level uncertainty, hidden dependence between module errors can produce substantially different downstream uncertainty estimates, collision risk, and robot behavior. Accounting for joint dependence restores downstream calibration, while conservative uncertainty bounds improve robustness at the cost of reduced efficiency. These results highlight the need for uncertainty interfaces that account for dependence or provide direct system-level calibration rather than relying on marginal calibration alone.

\clearpage
\onecolumn
\appendices
\refstepcounter{figure}\label{fig:sensitivity}
\begin{center}
\includegraphics[width=0.97\textwidth]{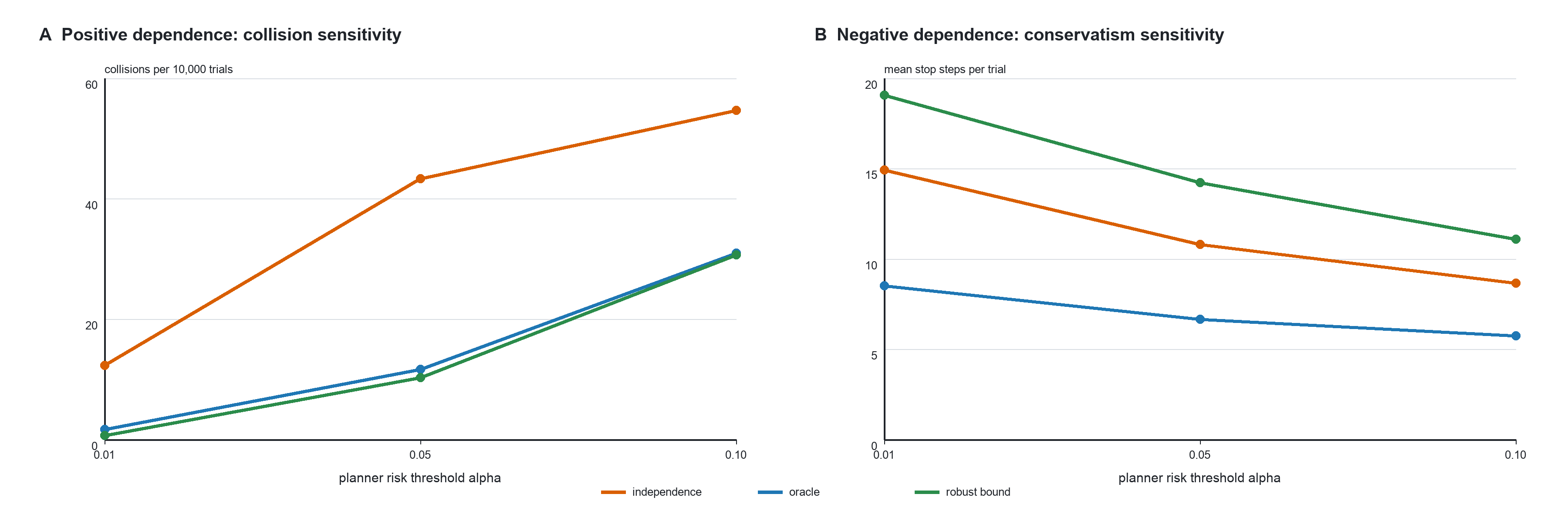}
\par\vspace{2pt}
{\small\textbf{Fig. \thefigure.} Planner-threshold sensitivity under extreme dependence. The positive-dependence collision ordering and negative-dependence stopping ordering persist at all tested thresholds. The robust bound follows the oracle under positive dependence but stops most under negative dependence. Each point uses 30,000 matched trials.}
\end{center}
\vspace{4pt}
\raggedbottom

\section{Planner-Threshold Sensitivity}\label{app:sensitivity}
At $\rho=+0.9$, collisions per 10,000 trials for independence versus oracle are 12.3 versus 1.7 at $\alpha=.01$, 43.3 versus 11.7 at $\alpha=.05$, and 54.7 versus 31.0 at $\alpha=.10$. The robust bound gives 0.7, 10.3, and 30.7, respectively. At $\rho=-0.9$, mean stop steps for independence versus oracle are 14.92 versus 8.52, 10.81 versus 6.65, and 8.67 versus 5.75. The robust values are 19.08, 14.23, and 11.10. The exact values differ slightly from Table~\ref{tab:results} because the sensitivity suite uses an independent fixed seed; conclusions are unchanged.

\section{Metric Definitions and Reproducibility}
For future error $e_i$ and reported variance $s_i^2$, we use
\begin{equation}
\mathrm{NLL}=\frac{1}{2N}\sum_i\left[\log(2\pi s_i^2)+e_i^2/s_i^2\right].
\end{equation}
For predicted conflict risk $q_i$ and event $c_i\in\{0,1\}$, the Brier score is $N^{-1}\sum_i(q_i-c_i)^2$. ECE partitions $[0,1]$ into 15 equal-width bins and averages the absolute difference between mean risk and event frequency, weighted by bin occupancy. Coverage and proper scores are primary; ECE is descriptive.

All experiments were executed on CPU using NumPy with fixed seed 20260719. Tables and figures were generated from the same experiment outputs, and the reported numerical values were checked for consistency.

\end{document}